\documentclass{article}
\usepackage{spconf,amsmath,graphicx}
\usepackage{amsmath}
\usepackage{enumerate}
\usepackage{subfigure}

\title{Deep learning with spatiotemporal consistency for nerve segmentation in ultrasound images}
\name{Adel Hafiane, Pierre Vieyres and Alain Delbos}
\address{INSA Centre Val de Loire, Laboratoire PRISME EA 4229, Bourges, France.\\
Universit\'{e} d Orl\'{e}ans, Laboratoire PRISME EA 4229, Bourges, France. \\
Clinique Medipole Garonne CS 13624, Toulouse, France.}
\begin{document}
%
\maketitle
\begin{abstract}
	Ultrasound-Guided Regional Anesthesia (UGRA) has been gaining importance in the last few years, offering numerous advantages over alternative methods of nerve localization (neurostimulation or paraesthesia). However, nerve detection is one of the  most tasks that anaesthetists can encounter in the UGRA procedure. Computer aided system that can detect automatically region of nerve, would help practitioner to concentrate more in anaesthetic delivery. In this paper we propose a new method based on  deep learning combined with spatiotemporal information to robustly segment the nerve region. The proposed method is based on two phases, localisation and segmentation. The first phase, consists in using convolutional neural network combined with spatial and temporal consistency to detect the nerve zone. The second phase utilises active contour model to delineate the region of interest. Obtained results show the validity of the proposed approach and its robustness.   
\end{abstract}
\section{Introduction}
\label{sec:intro}
The introduction of Ultrasound Guided Regional Anaesthesia (UGRA) 
has rapidly become popular for performing regional anaesthesia blocks. However, this procedure requires great concentration 
by the operator because of many simultaneous tasks, such as nerve and needle localization, and steady probe positioning in order to maintain the nerve and the needle in the observation plane~\cite{Marhofer2007},\cite{Carlos2012}. Performing the UGRA needs a long learning process and years of practice in the operating room. Computer aided system that can detect automatically region of interest ROI, would help practitioner to concentrate more in anaesthetic delivery.

Although there has been an extensive development of detection and segmentation algorithms for medical ultrasound (US) images \cite{Noble2006},\cite{Cheng10},\cite{Afsaneh13}, it is still an open problem especially for regional anesthesia. 
So far, very little attention has been paid to the nerve detection. 
We recently demonstrated the possibility of detecting and segmenting the sciatic nerve structure~\cite{Hafiane14}, the method is based on Monogenic signal and probabilistic active contour approaches. In~\cite{Hadjerci14} authors proposed a descriptor based on the combination of median 
binary pattern and Gabor filter to characterize and classify median nerve tissues. A machine learning framework was also proposed to enable robust detection of the median nerve~\cite{Hadjerci16}. Recent work addresses such a problem, by proposing assistive system that detect vessels and nerve region allowing path planning for needle~\cite{Hadjerci16b}. Despite the promising results obtained, still the topic requires further development and investigation.

Recently, the deep learning approach has emerged as a powerful approach to address many problems in machine learning and computer vision fields \cite{Hinton06},\cite{Bengio06} \cite{Bengio13}, \cite{Lecun2015}. Among deep learning methods, the Convolutional Neural Network (CNN) has been successfully applied  for various computer vision tasks, 
including objet recognition, region of interest (ROI) detection, segmentation and so on~\cite{Krizhevsky12}, \cite{Simonyan14}, \cite{Girshick14}, \cite{Long2015}.   
Deep learning is also gaining popularity in medical image segmentation and classification with promising results on various applications~\cite{Kooi2017},\cite{Tuan2017}.  

However, this approach is still limited to static images,
very little attention has been paid to dynamic information
generated by the motion of the probe~\cite{Hadjerci_ICIP16}. 
The purpose of this paper is to explore 
this type of information, motivated by the manner that human expert uses such information. 
In order to visualize region of interest (i.e. nerves, veins, arteries,..), clinician scan with the transducer a given location on the patient's body. Through this scanning process, he uses
dynamic information to increase confidence for nerve localisation in US images. Hence, it is interesting to use such kind of information to increase the robustness of detection and segmentation tasks.

In this paper we propose a new method based on convolutional neural network and spatiotemporal consistency to segment efficiently the nerve region. Indeed, CNN architecture is not sufficient to robustly locate the nerve region. Due to the noise and different artefacts, CNN may generate non negligible rate of false positives among the detected ROIs. In order to reduce this rate, we use spatial and temporal consistency to eliminate the false positives.  If a given position has a majority intersection of ROIs and steady with the same in time (through several US slices) the ROI is likely to be consistent and not a noisy one.
As result, the region of interest is considered as true positive. In final phase, we use active contours based on phase and probabilistic approach ~\cite{Hafiane14} to delineate the nerve contours.

The structure of this paper is organized as follows. In section 2, we present the method of nerve detection. Section 3 provides validation and evaluation of the proposed approach, followed by conclusion in Section 4. 
\begin{figure*}[ht]
	\centering
	\includegraphics[width=0.98\linewidth]{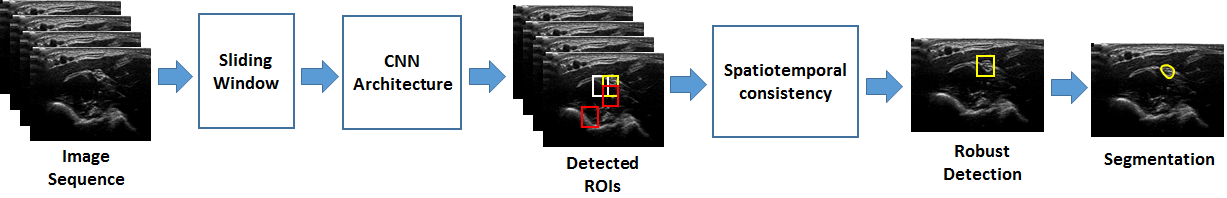}
	\caption{The flowchart of the proposed method}
	\label{fig:framework}
\end{figure*}
\section{Method} 
In order to segment nerve structures, we need to detect and locate them first. 
The localization procedure is  mainly based on two phases, 
the first one uses spatial coherence to keep selected ROIs with high probability measure. In the second phase it is assumed that the nerve chracteristics should be stable 
during certain period, when clinician steadly maintain the ultrasound probe. After localisation, the segmentation is much easier since the initial contours are near the target region. 
Figure~\ref{fig:framework} presents the general framework of the method. 
\subsection{ROIs detection with Convolutional neural networks}
Convolutional neural networks (CNNs) are one of the most effective deep learning approach, 
it is based on multilayer neural network. Generally, it uses three main neural layers: 
convolutional layers, pooling layers, and fully connected layers. 
The three main layers are deployed in cascading manner to perform the learning process. Typical convolutional layer consists in a 2D convolution operation with a set of kernels, which can be considered as a filtering process that generates a feature maps. The pooling layer performs downsampling operation on the filtered images, using maximum or average value of $2\times 2$ neighbourhood pixels. 
The fully connected layer follows the last operation of the pooling layer, 
it performs as the classical multi-layer neural network and provides a softmax output.

For a given input training set, CNN function is optimized to learn the best representation of target regions. Let $\mathcal{F}(x)$ be a learned function from CNN and $x$ as an input patch from US image. Given $x$, $\mathcal{F}$ predict the probability of $K$ classes
according to the learned weights $W$. Softmax function is used in CNN output layer to generate the probability $P(y=k|x,W) \in [0,1], k=1,2,\cdots,K$ such that:
\begin{equation}
P(y=k|x,W) = {\mathrm softmax}(\mathcal{F}(x))
\end{equation}
\begin{equation}
\hat{y} = \underset{k} {\mathrm {argmax}}\{P(y=k|x,W)\}
\end{equation}
$\hat{y}$ corresponds to class prediction, that takes the maximum probability of classes. 
However, with this approach, $\hat{y}$ may yield weak predictions; when 
probabilities of $K$ classes are slightly different from each other.
For ultrasound images, this could generate a large number of false positives since
images are corrupted with noise. For this reason we keep  $\hat{y}$ with high probability,
that is:
\begin{equation}
\label{Eq:ytilde}
\tilde{y} =  \underset{k} {\mathrm {argmax}}\{P(y=k|x,W)> \frac{\alpha}{K}\}
\end{equation}
where $\alpha$ is a parameter, set empirically, such that $\alpha<K$, for our experiments
$K=2$ and $\alpha=1.8$.

\subsection{Spatiotemporal consistency}
During the UGRA procedure practitioners use three basic motions of probe 
while scanning the patient; longitudinal sliding,  rotation and tilting. The probe position and motion affects the visual aspect of tissues in US images, 
anaesthetists adjust and stabilize the probe for best nerve visualisation.
They use back and forth motion of the probe on specific 
human body zone, to  generate dynamic information with consistent 
characteristics of certain anatomic structures such as nerves. 
Similarly, we explore dynamic information to reinforce ROIs detection.  
Indeed, detecting the nerve region in one US frame is not sufficient, due to variability on
its visual aspect on different frames. To robustly locate this region, it is more interesting to include temporal coherence over successive frames.

For this end, a sliding window scan the image and classify each location
with method described in previous section. Using Equation~\ref{Eq:ytilde}, we obtain several ROIs. Even though, $\tilde{y}$ yields ROIs with high probability, 
it is still a weak approach. In order to increase further the confidence in nerve localisation, we take into account the overlap between detected ROIs. A position that have several overlapped ROIs is considered as a valid candidate to be followed in US frames, for temporal consistency measure. Let $B_{i}^{t}$, ($i=1, 2,\cdots N$, $t=1, 2,\cdots, M$)  a $i^{th}$ candidate ROI in the frame $t$  to be identified as nerve zone or something else.  
The proposed scheme is formulated as: 
\begin{equation}
S_k^t=\{ B_i(x,y), k |r_k=\bigcap_i B_i(x,y) > 50\% , r_k > T\}
\end{equation}
where $r_k$ is the number of blocks (ROIs) that spatially overlap at least 50\%.
$S_k$ is a set of  ROIs, which satisfy the condition $r_k>T$.  $(x,y)$ are the coordinates of ROI in the image. $T$ is a threshold that determines the number of overlapped blocks. In the present experiments, $T=3$.

Temporal coherence is measured by the number of ROIs that are consistent in position $(x,y)$ over $M$ frames. It is given by:
\begin{equation}
f_k = \sum_t^M | S_k^t | 
\end{equation}
where $| S_k^t |$ is the cardinal of the set $S_k$ in the $t^{th}$ frame.
Finally, robust nerve localisation is determined by the maximum spatiotemporal consistency.  
\begin{equation}
g(f_k) = \underset{k} {\mathrm {argmax}} \{f_k\}
\end{equation}

Note that, the localisation decision is applied to the final frame, since the model was built
with frames history.

\subsection{Active model for segmentation}
After localisation we need to delineate the nerve contours.
For that purpose, we use phase based probabilistic active contour (PGVF)~\cite{Hafiane14}, since it provides better results for nerve segmentation compared to classical methods. The bounding box of the localization is used as an initial contour. The PGVF function is based on the combination of the probabilistic learning approach, with the local phase information. It modifies the external energy equation of the original GVF (Gradient Vector Flow).

\section{Experiment and results}
\subsection{dataset}
The dataset was obtained in real condition at Medipole Garonne hospital at Toulouse in France. Ultrasound machine dedicated to the regional anaesthesia was used to acquire the dataset. 
A linear transducer probe with 5-12 MHz was utilized, the acoustic beam generated by the transducer produces a series of pulse echo lines that generates transverse section images of anatomic structures. 
To visualize the median nerve block, anaesthetists scan a given zone on the forearm using 
four basic movements on the probe,  translation, alignment, rotation and tilt.
Therefore, a sequence of images is generated during the scanning procedure. The
image sequence was saved as a video.

In this study we used dataset that contains ten videos of median nerve corresponding to ten patients. The dataset was annotated by regional anesthesia experts, 
providing ground truth of the nerve region in the images. 
For learning and testing phases, a sequence of frames is extracted from each video, we have in average 500 images per video, the size of each image is $600\times300$. To feed the CNN model, we used patches with size of $64\times 64$ that represent positive class (nerve region), and negative class (non nerve).

The first step consists in learning features and establish classifier model with CNN architecture. The optimization of the network were performed using stochastic gradient descent (SGD).
We used a 0.5 dropout rate on the fully connected layer during the training stage for regularization. We employed the CNN architecture with 3 convolutional layers using $3\times 3$ kernels followed by $2\times 2$ max-pooling and ReLU activation function, then a fully connected layer of 128 input was added. The last layer consists in softmax function that generates a probability of the two classes. 
\subsection{Performance evaluation}
First we examine the classification ability of the proposed method for nerve localisation in ultrasound images. For that purpose, we adopted cross validation  with $10-fold$ procedure for performance evaluation. In second stage, we evaluate the nerve segmentation after its localisation using Dice and Hausdorff metrics.

\subsubsection{Qualitative evaluation}
\begin{figure*}[t]
	\centering
	\subfigure[]{\includegraphics[width=0.3\linewidth,height=3cm]{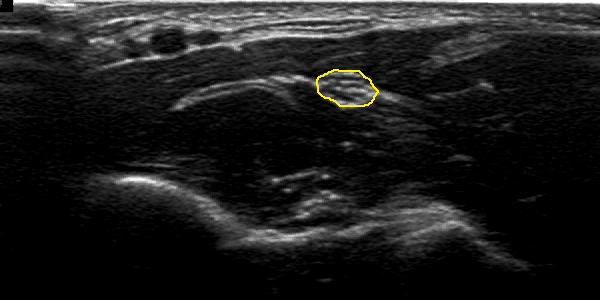}}
	\subfigure[]{\includegraphics[width=0.3\linewidth,height=3cm]{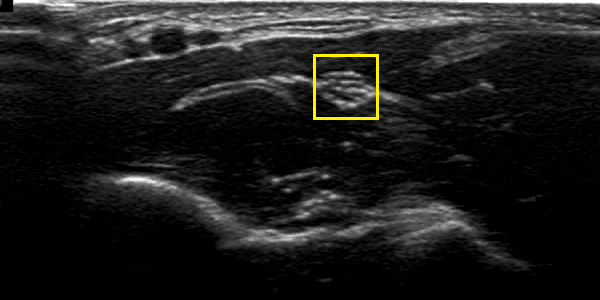}}
	\subfigure[]{\includegraphics[width=0.3\linewidth,height=3cm]{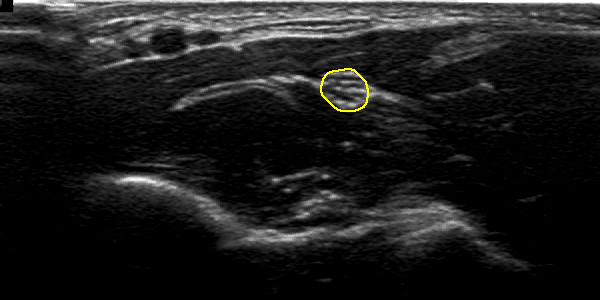}}
	\\
	\subfigure[]{\includegraphics[width=0.3\linewidth,height=3cm]{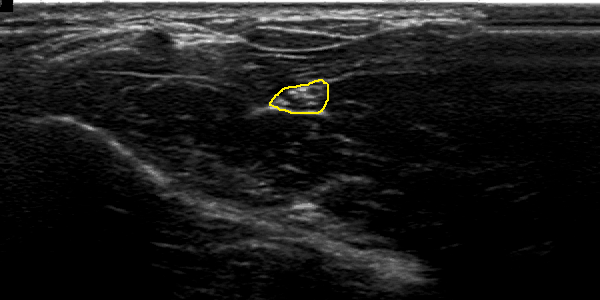}}
	\subfigure[]{\includegraphics[width=0.3\linewidth,height=3cm]{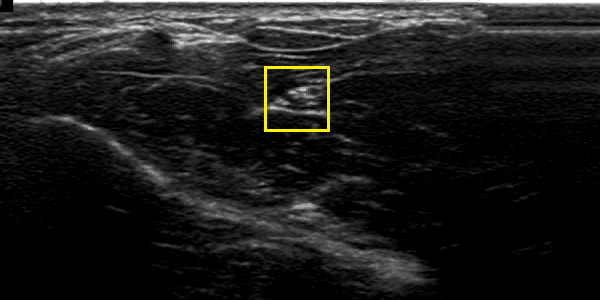}}
	\subfigure[]{\includegraphics[width=0.3\linewidth,height=3cm]{MEDIAN10_21_GT}}
	\caption{An example of the obtained using deep learning with spatio temporal consistency. Images show the median nerve of two different patients. Left column (a) and (d), each image is the ground truth of the median nerve of each patient. The middle column (b) and (e), represents the localisation result using CNN and spatiotemporal constraint. Right column (c) and (f), shows the segmentation result using PGVF algorithm}
	\label{fig:result1}
\end{figure*}

Figure~\ref{fig:result1} illustrates and example of results in two US images from two different patients. Figures~\ref{fig:result1} (b) and (e) shows 
the localisation obtained with CNN and spatiotemporal consistency. 
The nerve segmentation by PGVF is depicted in figures~\ref{fig:result1} (c) and (f).  
One can observe that the automatic segmentation is very close to the one obtained form  human experts figures~\ref{fig:result1} (a) and (d).

\subsubsection{Quantitative evaluation}
We compare our proposed method with similar approaches based on Support Vector Machine (SVM) classifier.
We consider a true positive localization if the detected ROI overlap the ground truth ROI at least 50\%, otherwise the detected ROI is considered as false positive. Results present average performance on 5000 images. Table~\ref{tab:tab1} summarize the results obtained with cross validation procedure. As we can observe CNN do better than SVM with feature selection.
SVM combined with spatiotemporal coherence~\cite{Hadjerci_ICIP16} increase performances further more, but CNN combined with
spatiotemporal consistency provides best results.  

\begin{table*}[htbp]
	\centering
	\begin{tabular}{@{} cccc @{}}
		\hline
		Method & precision & recall & F-score \\ 
		\hline
		Our method 	 					 & {\bf $0.89 \pm 0.17$}& {\bf $0.96 \pm 0.21$}& {\bf $0.96 \pm 0.25$} \\ 
		CNN 	     					 & $0.87 \pm 0.15$ & $0.94 \pm 0.23$ & $0.91 \pm 0.24$ \\
		SVM + feature selection \cite{Hadjerci16}  & $0.81 \pm 0.19$ & $0.93 \pm 0.28$  & $0.88 \pm 0.52$  \\ 
		SVM +temporal constraint \cite{Hadjerci_ICIP16} & $0.85 \pm 0.17$ & $0.95 \pm 0.21$  & $0.92 \pm 0.33$  \\ 
		\hline
	\end{tabular}
	\caption{Comparison between nerve localisation methods}
	\label{tab:tab1}
\end{table*}

Once the nerve zone is detected, the next step utilizes active contour approach to detect nerve borders.
This segmentation approach is evaluated according to the ground truth. Table~\ref{tab:tab2} presents
the average performance of final segmentation. 
Overall, these results indicate that the deep learning combined wit spatiotemporal constraint increase detection and segmentation performances.

\begin{table}[htbp]
	\centering
	\begin{tabular}{@{} ccc @{}}
		\hline
		Method & Dice metric & Hausdorff metric \\ 
		\hline
		Localisation + PGVF~\cite{Hafiane14} 	 & {\bf $0.85 \pm 0.15$} &  {\bf $10.72 \pm 3.91$}  \\ 
		\hline
	\end{tabular}
	\caption{Evaluation of active model (PGVF) segmentation using the detected ROI as initial contour.}
	\label{tab:tab2}
\end{table}

\section{Conclusion}
In this paper we presented a new method that combine Deep learning approach and spatiotemporal concept. The method is applied successfully to segment median nerve structure in ultrasound images. 
To reduce the false positive rate, we combined spatial and dynamic information together with the CNN classifier. Nerve localisation worked better than CNN alone or SVM with spatiotemporal coherence. The comparative study showed the effectiveness of the proposed scheme,
achieving 96\% of F-score. Despite this promising results reported so far, there is room for further development. For instance, targeting other types of nerves, increasing dataset for learning and test. 

\bibliographystyle{IEEEbib}
\bibliography{biblio}

\end{document}